\documentclass[10pt,twocolumn]{article}
\usepackage[margin=1in]{geometry}
\usepackage[utf8]{inputenc}
\usepackage[T1]{fontenc}
\usepackage{lmodern}
\usepackage{microtype}
\usepackage{amsmath,amssymb}
\usepackage{graphicx}
\usepackage{caption}
\usepackage{subcaption}
\usepackage{float}
\usepackage{booktabs}
\usepackage{url}
\usepackage{fvextra}
\usepackage{enumitem}
\usepackage{xcolor}
\usepackage{hyperref}
\hypersetup{colorlinks=true,linkcolor=blue,citecolor=blue,urlcolor=blue,filecolor=blue}
\graphicspath{{../extracted/images/}}
\DefineVerbatimEnvironment{PromptVerbatim}{Verbatim}{
  breaklines=true,
  breakanywhere=true,
  fontsize=\footnotesize,
  frame=single,
  framerule=0.5pt,
  rulecolor=\color{gray},
  framesep=6pt
}

\title{PhysicsSolutionAgent: Towards Multimodal Explanations for Numerical Physics Problem Solving}

\author{Aditya Thole, Anmol Agrawal, Arnav Ramamoorthy, Dhruv Kumar \\
\small\texttt{\{f20220374, f20221313, f20220007, dhruv.kumar\}@pilani.bits-pilani.ac.in} \\
\small\url{https://github.com/f20220374/PhysicsSolutionAgent}}

\date{}

\begin{document}
\maketitle

% Fig 1 at the top spanning both columns
\begin{figure*}[t]
  \centering
  \includegraphics[width=0.92\textwidth]{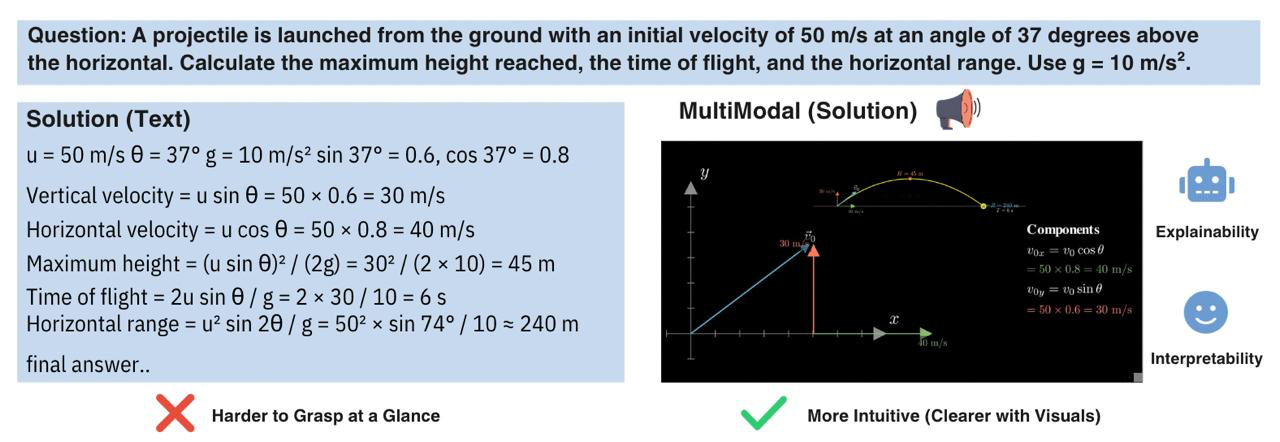}
  \caption{Comparison between a traditional text-based physics solution and the multimodal solution generated by PhysicsSolutionAgent, showing how visual explanations improve clarity and intuition.}
  \label{fig:comparison}
\end{figure*}

\begin{abstract}
Explaining numerical physics problems often needs more than text-based solutions; clear visual reasoning can make the concepts much easier to understand. While large language models (LLMs) solve many physics questions well in text form, their ability to generate long, high-quality visual explanations is still not fully understood. In this work, we introduce PhysicsSolutionAgent (PSA), an agent that generates up to 6-minute physics-problem explanation videos using Manim animations. To evaluate these videos, we created a pipeline that runs automated checks on 15 parameters and also uses a vision-language model (VLM) to give feedback and improve video quality. We tested the system on 32 videos covering numerical and theoretical physics questions. Our results show clear differences in video quality depending on the difficulty of the numerical problem and whether the question is theoretical. Using GPT-5-mini, our agent achieved a 100\% video-completion rate and an average automated score of 3.8/5. However, human interpretation and qualitative analysis reveal several minor and major issues, both in the visual layout of elements and in how the AI evaluates and interprets images during feedback. These findings highlight key limitations in the agent’s ability to write reliable Manim code and the broader challenges in multimodal reasoning for numerical physics explanations, underscoring the need for better visual understanding and evaluation methods in future systems.
\end{abstract}

\section{Introduction}
Agentic AI systems that leverage Large Language Models (LLMs) are rapidly advancing the frontier of automated content creation. In the educational domain, a significant milestone is the development of frameworks that can generate multimodal video explanations for complex STEM topics. The ``TheoremExplainAgent'' \cite{ku2025theoremexplain} established a powerful baseline, demonstrating that a planner-coder agent architecture can produce long-form, animated videos for a wide range of scientific theorems. This approach holds immense promise for making education more accessible and scalable. However, to move from promising demonstrations to robust, production-quality systems, a deeper, more quantitative understanding of their performance is required.

Key questions remain: How consistently do these agents perform across different types of tasks, such as explaining a broad concept versus solving a specific numerical problem? What are the most common failure modes or quality issues, and can they be systematically identified and addressed?

To investigate these questions, we present a specialized agentic framework architectured to generate high-fidelity instructional videos for physics. Our pipeline employs planner and coding agents powered by GPT-5 mini to drive the Manim animation engine. Our architecture introduces several key enhancements for robustness and quality: a Retrieval-Augmented Generation (RAG) module that provides the coding agent with relevant Manim documentation, a multi-attempt error-correction loop that enables the agent to debug its own code, and a screenshot-driven feedback loop where a Vision Language Model (VLM) analyzes rendered frames and directs the agent to make targeted visual improvements.

We introduce a novel automated evaluation system that captures 15 distinct metrics for every generated video, allowing for a detailed, data-driven analysis of the agent's performance. Our evaluation, conducted on prompts covering both conceptual theorems (e.g., ``Explain Snell's Law'') and multi-step numerical problems (e.g., ``Calculate the final speed of a block on an incline''), yields two primary findings. First, our specialized agent demonstrates a high degree of proficiency across both categories, achieving an average weighted quality score of 3.82/5 for conceptual videos and a nearly identical 3.80/5 for numerical problems. This consistent, high-level performance, especially on the visually and logically complex numerical tasks, validates the effectiveness of combining a specialized architecture with a powerful model like GPT-5 mini.

Second, our granular, automated evaluation successfully identifies a consistent pattern of minor, yet important, quality issues. Across both video types, the most frequent weaknesses were not fundamental errors in physics but rather ``minor redundancy in repeated visualization text'' and opportunities to ``condense repeated explanations.'' These findings provide a clear, actionable insight: the areas for improving agentic educational video generation lies in refining the agents' ability to produce more concise and polished narrative and visual content.

Our contributions are:
\begin{itemize}[leftmargin=*]
  \item We present a specialized agentic framework enhanced with RAG and a visual feedback loop that achieves consistently high performance on both conceptual and numerical physics video generation.
  \item We provide a detailed, data-driven analysis based on a comprehensive, 15 metric automated evaluation, offering a quantitative snapshot of the current state of the art.
  \item We identify and characterize minor textual and visual redundancy as a key, recurring area for improvement in agent-generated instructional content, paving the way for more refined and professional video outputs.
\end{itemize}

\section{Related works}
Our work is situated at the intersection of several key research areas: automated multimodal content generation, agentic AI systems for scientific tasks, and the use of code-driven animation tools for STEM education.

\subsection{Automated Generation of Multimodal Explanations}
Bridging the gap between dense textual information and intuitive visual explanations is an emerging research direction, grounded in multimedia learning principles \cite{mayer2009multimedia}. Most current approaches focus on text-to-image or text-to-static-diagram generation. However, generating dynamic, structured animations for educational purposes presents unique challenges in temporal consistency, pedagogical structuring, and the accurate representation of symbolic information like mathematical formulas.

The most closely related and foundational work is TheoremExplainAgent \cite{ku2025theoremexplain}. This work introduced a novel agentic framework for generating long-form video explanations of STEM theorems using the Manim animation engine \cite{manim2024}, originally developed for the popular 3Blue1Brown educational channel \cite{sanderson2024}. It employs a planner agent to create a storyboard and a coding agent to generate the corresponding Manim Python script. Ku et al. \cite{ku2025theoremexplain} also proposed the TheoremExplainBench (TEB), a standardized benchmark with five automated evaluation metrics, which we adopt for our evaluation to ensure a direct and fair comparison.

Another relevant system is Manimator \cite{p2025manimator}, which also uses a multi-stage LLM pipeline to transform natural language prompts or research papers into Manim animations. Manimator's findings reinforce the viability of the planner-coder pipeline and highlight that the choice of a code-specialized LLM (DeepSeek-V3 \cite{liu2024deepseek}) can significantly improve visual quality, particularly the ``Element Layout'' metric on the TEB benchmark. Our work builds on this agentic paradigm but introduces a novel visual feedback loop, where a VLM analyzes rendered frames to iteratively refine the output, directly addressing the ``minor visual layout inaccuracies'' identified as a key challenge in prior work.

\subsection{LLMs for code-driven animations}
The Manim library is a powerful tool for creating precise, programmatically generated animations, but its complexity represents a significant bottleneck for educators without programming expertise \cite{helbling2023manimml}. Automating Manim code generation with LLMs \cite{chen2021codex} is therefore a critical component of scalable video creation.

This task, however, is non-trivial. Ku et al. \cite{ku2025theoremexplain} identified ``Manim code hallucinations'' and ``LaTeX rendering errors'' as primary failure categories for their system. Our work addresses this challenge through a multi-pronged strategy. First, we employ a Retrieval-Augmented Generation (RAG) \cite{lewis2020rag} system where our coding agent queries a ChromaDB vector database of Manim documentation before writing code. This grounds the LLM, reducing the likelihood of hallucinating non-existent functions or incorrect parameters. Second, we implement a robust, multi-attempt error-correction loop, allowing the agent to analyze stack traces and debug its own code, a feature also present in TheoremExplainAgent. Finally, our use of GPT-5 mini, provides an additional advantage in generating complex and syntactically correct Manim scripts.

\subsection{Agentic Frameworks for Scientific Tasks}
The use of LLM-based agents that can plan, act, and refine their work is a rapidly growing field. In the domain of scientific visualization, systems like PlotGen \cite{goswami2025plotgen} have explored using multimodal feedback for iterative refinement. Our work introduces a novel application of this concept to video generation. While the error-correction loop in our system (and in TheoremExplainAgent) represents a form of feedback, it is reactive to programming errors. Our screenshot-driven feedback loop is a more sophisticated mechanism. After an initial video is rendered, a VLM analyzes the visual output for aesthetic and pedagogical issues like element overlap, poor text readability, or content being cut off. These qualitative visual critiques are then translated back into a new prompt for the coding agent to refine the Manim code. This positions our framework as a more advanced agent that learns not just from execution failures, but from the quality of its own multimodal output, pushing the state of the art towards higher-fidelity, more polished educational content.

% Fig 2 spanning columns near methodology
\begin{figure*}[t]
  \centering
  \includegraphics[width=0.95\textwidth]{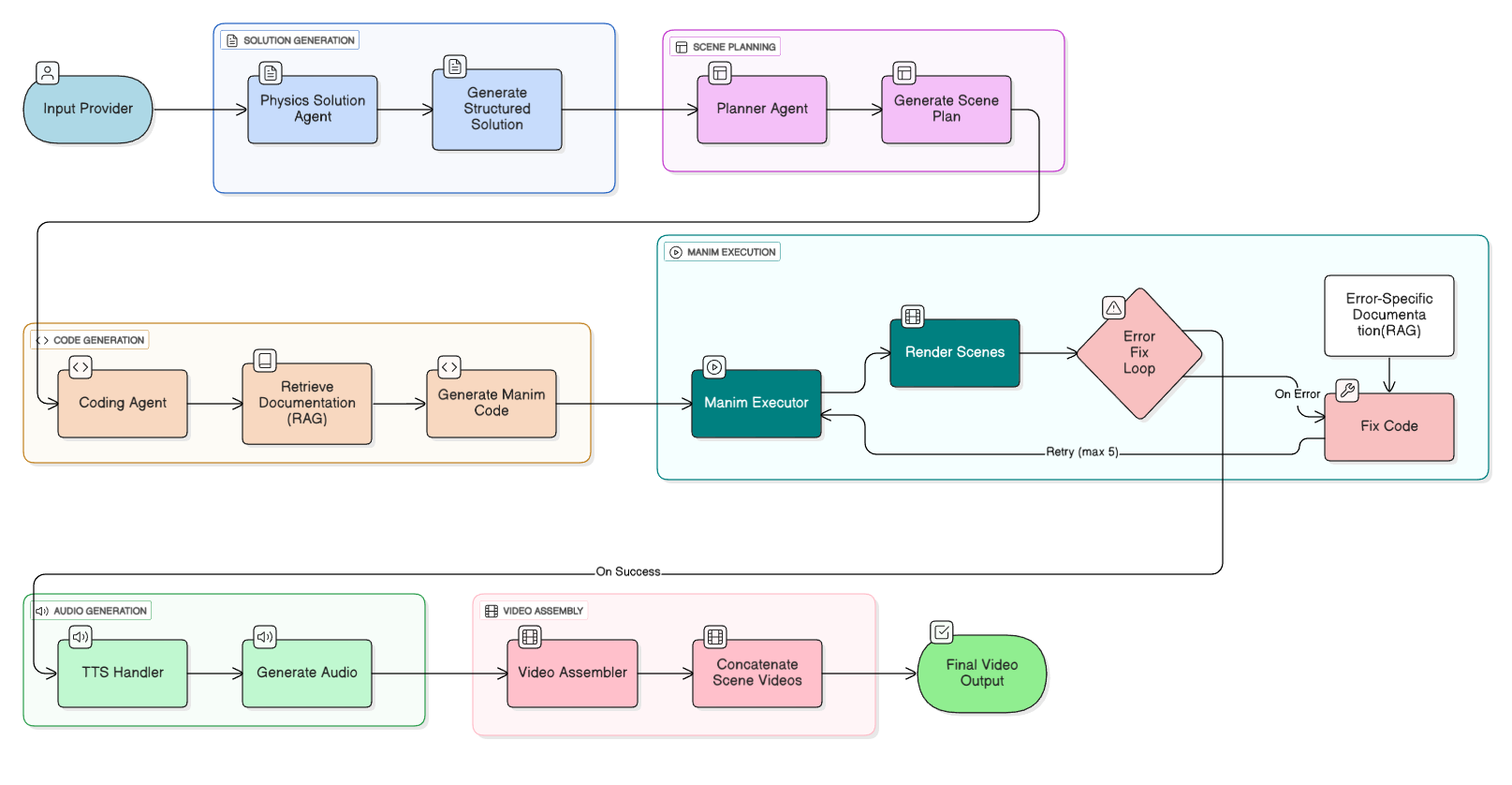}
  \caption{The end-to-end video generation pipeline. Specialized agents collaborate to solve the physics problem, plan the visual narrative, and generate code.}
  \label{fig:pipeline}
\end{figure*}

\section{Methodology}
We introduce a specialized agentic framework designed to automate the generation of high-fidelity instructional videos for physics, moving from a textual prompt to a final animated video. Our methodology is structured as a multi-stage pipeline, as illustrated in Figure~\ref{fig:pipeline}. The core of our system is a two-agent architecture, a Planner Agent and a Coding Agent, enhanced with a Retrieval-Augmented Generation (RAG) system for robust code synthesis and a screenshot-driven feedback loop for visual refinement.

\subsection{Stage 1: Solution Generation}
The pipeline begins with the PhysicsSolutionAgent. Given a user-provided physics question as a string input, this agent utilizes an LLM to generate a comprehensive, step-by-step solution following chain-of-thought reasoning principles \cite{wei2022cot}. The output is a structured JSON object containing a detailed breakdown of the problem, including the relevant theoretical concepts, mathematical derivations, and final conclusions. This structured solution serves as the ground-truth knowledge base for all subsequent stages.

\subsection{Stage 2: Scene Planning}
The structured solution is then passed to the PlannerAgent. This agent acts as an instructional designer, analyzing the solution to create a pedagogically sound video plan. It deconstructs the explanation into a sequence of logical scenes. For each scene, it generates a title, a clear educational purpose, a detailed description of the visual elements, and a precise narration script. The output is a JSON array of scene objects that serves as the blueprint for the animation.

\subsection{Stage 3: RAG-Enhanced Code Generation}
The scene plan is executed by the CodingAgent, a specialized agent tasked with translating the visual and narrative specifications into executable Python code using the Manim library. To enhance the robustness and accuracy of the generated code, this agent is augmented with two key features: Retrieval-Augmented Generation (RAG) that grounds the LLM with relevant Manim documentation, and context learning via few-shot examples of high-quality, well-structured Manim code. The agent's final output is a complete Python script that includes all necessary Manim classes and integrates the narration using the manim-voiceover plugin with the Kokoro TTS service \cite{kokoro2024}.

\subsection{Stage 4: Execution with Error-Correction Loop}
The generated Manim script is executed on a per-scene basis using a subprocess. This stage includes a critical error-correction loop designed to handle the inherent fallibility of LLM-generated code. If a Manim execution fails: (1) stderr output containing the Python traceback is captured; (2) this error message, along with the faulty code and the original scene plan, is fed back to the CodingAgent; (3) the agent analyzes the error and generates a corrected version of the code. This loop repeats up to a maximum of five attempts, allowing the system to autonomously resolve a significant portion of common programming errors, such as LaTeX syntax issues and incorrect function calls.

\subsection{Stage 5: Screenshot-Driven Visual Refinement}
After an initial, error-free version of the complete video is assembled, our framework executes a one-shot visual refinement loop. Static screenshots are captured from the start, middle, and end of each rendered scene. A Vision Language Model (VLM) analyzes these frames against a detailed visual quality rubric, identifying issues such as element overlap, poor text readability, or content being cut off at the screen edges. The VLM's qualitative feedback is aggregated into actionable change instructions and passed to the CodingAgent, which then modifies the original Manim script to address the identified visual flaws. The improved script is re-executed (using the same error-correction loop) to produce a visually refined final set of video scenes, which are then assembled.

\section{Experimental Setup}

\begin{table*}[t!]
  \centering
  \caption{Mean Scores by Task Category (v2)}
  \label{tab:category}
  \begin{tabular}{lccc}
    \toprule
    Category & Mean OS (Std Dev) & Mean VQS (Std Dev) & Avg. Time (s) \\
    \midrule
    Easy   & 3.90 (0.44) & 3.42 (0.32) & 863.48 \\
    Medium & 3.79 (0.23) & 3.34 (0.20) & 929.06 \\
    Hard   & 3.72 (0.19) & 3.23 (0.19) & 1130.05 \\
    Theorem& 3.82 (0.31) & 3.38 (0.31) & 1066.11 \\
    \bottomrule
  \end{tabular}
\end{table*}

We conduct a comprehensive and automated evaluation to rigorously assess the performance of our specialized agentic framework. Our experimental setup is designed to measure the quality and robustness of the generated videos across different task types and to provide a clear, quantitative basis for our findings.

\subsection{Dataset and Task Categorization}
To analyze the agent's performance across varied complexities, we use a dataset of textual prompts divided into two primary categories: (1) Conceptual ``Theorem'' Problems requiring broad explanations (e.g., Archimedes’ Principle), and (2) Numerical Problems providing a specific, quantitative problem requiring a step-by-step mathematical solution. Numerical problems are further sub-categorized by difficulty: Easy, Medium, Hard. For each prompt, the system generates Version 1 and a visually-refined Version 2 outputs to measure the impact of the screenshot-driven feedback loop.

\subsection{Automated Evaluation Framework}
Our evaluation is performed by a fully automated LLM-as-a-Judge system \cite{pathak2025rubric}. It assesses each generated video and its intermediate artifacts against a multi-dimensional rubric, producing a final Score on a 0--5 scale. The overall score is a weighted combination of Solution Quality (5\%), Explanation Quality (10\%), Visual Quality (60\%), and Error Penalty (25\%). Visual Quality itself averages strict sub-metrics: Layout quality, Text readability, Equation rendering, Off-screen issues, and Scene-content alignment. We additionally log qualitative feedback: top issues and actionable recommendations.

\subsection{Data Logging and Qualitative Analysis}
For each generated video, we log all quantitative scores and supplementary data into a cumulative CSV, including per-stage timing and error counts. Qualitative feedback highlights recurrent issues like redundancy and verbosity, visuals/graphics issues, and LaTeX formatting problems.

\section{Results}

\subsection{Quantitative Performance by Task Difficulty}
We evaluated the final Version 2 videos across four problem categories. Table~\ref{tab:category} summarizes the mean Overall Score (OS), Visual Quality Score (VQS), and average execution times.

Key findings: High baseline performance (mean \~3.8/5), consistency across complexity (low std devs), and expected computational cost increases for Hard problems.

\subsection{Impact of Visual Refinement}
Comparing Version 1 vs Version 2 shows consistent, modest improvements: Layout Quality (3.64→3.66), Scene-Content Alignment (3.31→3.53), and Max Scene Visual Quality (3.98→4.05), indicating polishing effects of the feedback loop.

\subsection{Qualitative Analysis of Recurring Issues}
Automated analysis reveals the most prevalent issue is redundancy and verbosity (appearing 36 times), followed by visuals/graphics issues (18), and LaTeX formatting problems. Theorem questions had higher frequency of visual issues, likely due to abstract visualization challenges.

\subsection{Execution Time Analysis}
Hard problems averaged ~19 minutes per generation, ``Theorem'' showed higher variance, highlighting trade-offs for high-fidelity educational content.

\section{Limitations}
We limit the screenshot-driven refinement loop to a single iteration and only 3 screenshots/scene to avoid infinite loops and contain costs. Static screenshots miss temporal properties like animation smoothness and audio-visual synchronization. The most persistent issue is redundancy in narration and visuals; fine-tuning on concise scripts could help. RAG currently targets only Manim documentation; lack of an external physics knowledge base could still allow hallucinations on niche content. Complex problems have higher latency, unsuitable for real-time tutoring.

\section{Conclusion}
We presented a specialized agentic framework that bridges conceptual understanding and procedural mastery in physics education by orchestrating a Planner-Coder architecture powered by GPT-5 mini, augmented with RAG and screenshot-driven feedback. Across a diverse dataset, the system maintains consistent quality for both quantitative and conceptual tasks and benefits from multimodal reflection. Future work should emphasize conciseness and real-time constraints.

\section*{Acknowledgement}
We thank the Manim Community Developers for the open-source Manim library, and the authors of TheoremExplainAgent and Manimator for their pioneering work in agentic video generation which informed our evaluation and pipeline design.

\nocite{*}
\bibliographystyle{ieeetr}
\bibliography{references}

\onecolumn
\appendix
\section*{Appendix}

\begin{figure}[H]
  \centering
  \includegraphics[width=0.9\textwidth]{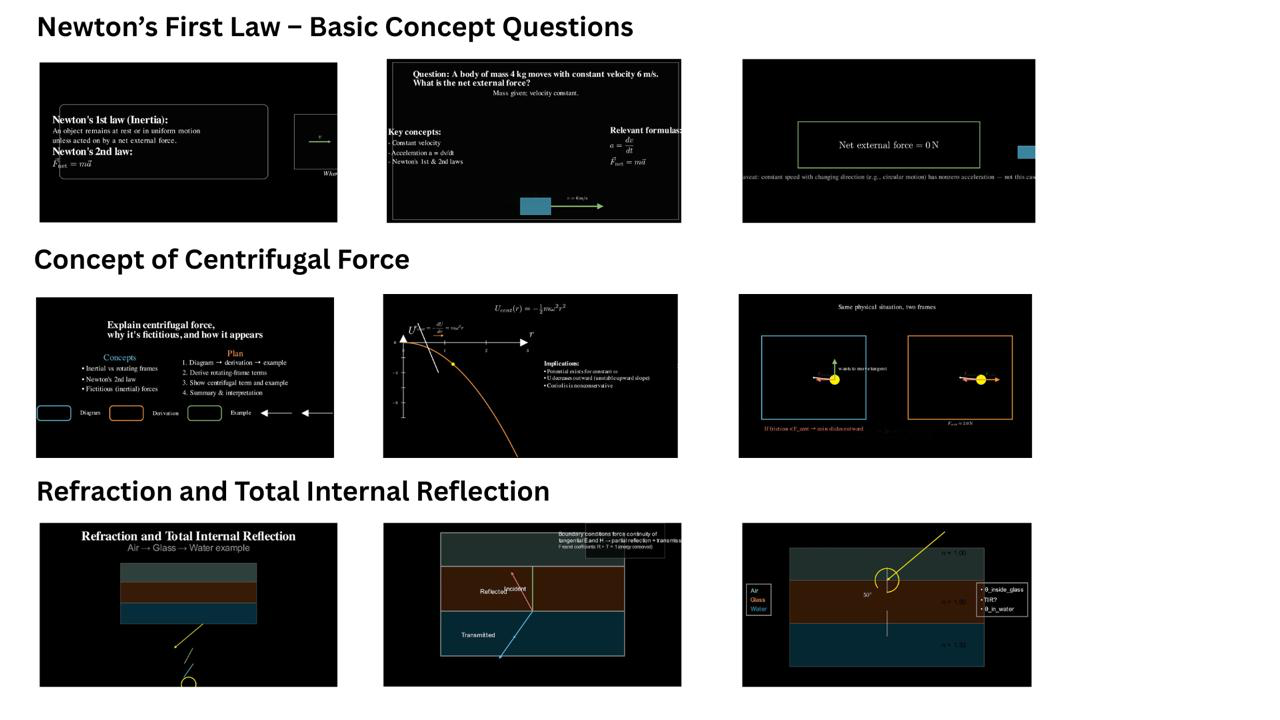}
  \caption{Examples of poor quality scenes.}
  \label{fig:poor}
\end{figure}

\begin{figure}[H]
  \centering
  \includegraphics[width=0.9\textwidth]{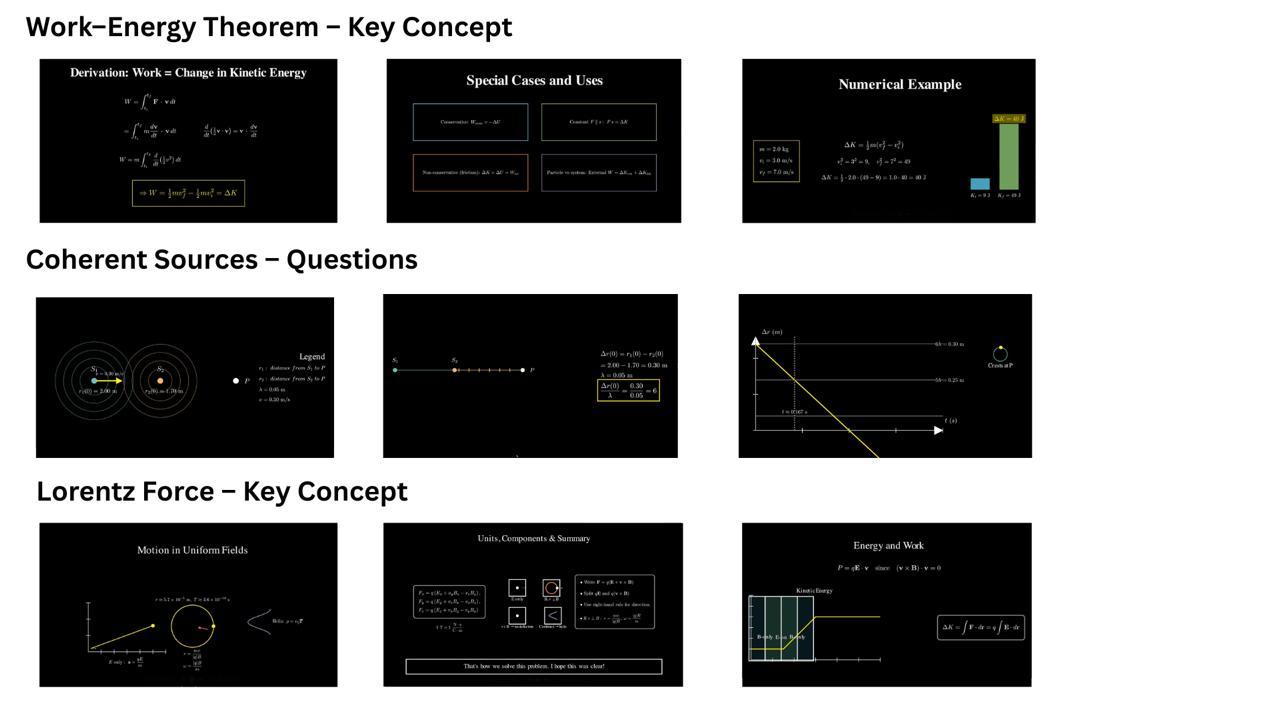}
  \caption{Examples of high quality scenes.}
  \label{fig:high}
\end{figure}

\vspace{8pt}

\subsection*{Prompt Templates}
\vspace{6pt}
\subsubsection*{1. PHYSICS SOLUTION PROMPT}
\vspace{4pt}
\begin{PromptVerbatim}
Agent: PhysicsSolutionAgent
Purpose: Solve JEE-level physics problems step-by-step
Output: Structured solution with equations, calculations, and visualization hints

You are an expert physics teacher specializing in JEE (Joint Entrance Examination)
level problems. Your task is to solve a given physics problem step by step,
showing clear reasoning, equations, calculations, and explanations suitable for
visualization.

PHYSICS QUESTION:
{question_text}

Please provide a complete solution following this structured format:

SOLUTION BEGIN:
[Step 1]
Approach/Concept: Identify the key physics concepts and principles needed.
Equation/Formula:       Write     the     relevant        equations/formulas.
Calculation: Show step-by-step calculations with units.
Explanation: Explain why this step is important and how it leads to the solution.
Visualization: Describe what should be visualized (diagrams, graphs, vectors, etc.).

[Step 2]
... continue with all necessary steps ...

SOLUTION END:

IMPORTANT REQUIREMENTS:
1. Break the solution into clear, logical steps
2. Start with conceptual understanding before calculations
3. Include all relevant formulas with proper notation
4. Show detailed calculations with units
5. Explain the physics behind each step
6. Include visualization suggestions that help understand the concepts
7. Focus on JEE-level rigor and accuracy
8. Ensure steps flow logically from given information to final answer

Example format for a projectile motion problem:
SOLUTION BEGIN:
[Step 1]
Approach/Concept: This is a 2D projectile motion problem where we need to decompose
motion into horizontal and vertical components.
Equation/Formula: v_x = v0*cos(theta), v_y = v0*sin(theta) - gt
Calculation: For initial velocity 10 m/s at 45°:
v_x = 10cos(45°) = 7.07 m/s
v_y = 10sin(45°) - g t
Explanation: The horizontal velocity remains constant due to no horizontal forces.
Visualization:     Show         velocity     vector decomposition with angle, draw x and y components.

[Step 2]
...continue with trajectory calculation, maximum height, range, etc...
SOLUTION END:

Remember to be rigorous and thorough, as this solution will be used to generate an
educational video explanation.
\end{PromptVerbatim}

\vspace{8pt}
\subsubsection*{2. SCENE PLANNING PROMPT}
\vspace{4pt}
\begin{PromptVerbatim}
Agent: PlannerAgent
Purpose: Break down solution into visual scenes for video production
Output: Scene-by-scene plan with titles, purposes, descriptions, layouts, narration

You are an expert in video production, instructional design, and JEE physics
education. Please design a high-quality video to explain the solution to a
physics problem.

Video Overview:
Question: {question}
Solution Steps: {solution}

Scene Breakdown:
Plan individual scenes that explain this physics solution. For each scene provide:
- Scene Title: Short, descriptive title (2-5 words).
- Scene Purpose: What physics concept or solution step does this scene explain?
- Scene Description: Detailed description of visual content, equations, diagrams.
- Scene Layout: Detailed description of the spatial layout concept. Consider
  equations, diagrams, vectors, graphs. Ensure minimum spacing between elements.
- Narration Script: CONCISE teacher-style narration (20-40 words max per scene).
  Like a teacher pointing to a board, guide attention to key elements without
  reading everything displayed.

Please generate the scene plan in this format:
SCENE PLAN BEGIN:
[Scene 1]
Title: Problem Statement and Approach
Purpose: To present the question and identify key physics concepts
Description: Display the question text, then highlight key physics concepts and
relevant formulas needed for solution.
Layout: Question at top, then split screen: left side shows list of relevant
concepts, right side shows key formulas.
Narration: "Here's our problem. Notice we're dealing with circular motion and
forces. We'll use Newton's second law and centripetal force concepts."

[Scene 2]
...continue with solution steps...
SCENE PLAN END:

CRITICAL NARRATION GUIDELINES:
1. Keep narration BRIEF (20-40 words per scene) - like a teacher highlighting key points
2. FIRST SCENE must start with a warm teacher introduction: "Hello! Today we will
     learn about..." or "Welcome students! Let's understand..."
3. LAST SCENE must end with encouraging conclusion: "I hope you understand the
    solution now!" or "That's how we solve it. I hope this was clear!"
4. DO NOT read equations or text displayed on screen
5. Point out what's IMPORTANT: "Notice the key term here" or "This step is crucial
    because..."
6. Guide attention: "Look at the force diagram" or "See how energy is conserved"
7. Explain WHY, not WHAT: "We use this formula because..." not "The formula is..."
8. Each scene's narration should match its visual content duration (3-8 seconds of speech)
9. Avoid long sentences - use short, clear phrases
10. Focus on insights and connections, not descriptions

VIDEO STRUCTURE GUIDELINES:
1. Start with clear problem statement
2. Break solution into logical steps (6-12 scenes total)
3. Emphasize visual representations (diagrams, graphs, vectors)
4. Highlight key equations and calculations
5. Include brief explanations of physics concepts
6. Show clear connections between steps
7. End with solution verification and key takeaways

Create scenes that help students understand both HOW to solve the problem and WHY
each step works. Remember: Visual content teaches, narration guides attention.
\end{PromptVerbatim}

\vspace{8pt}
\subsubsection*{3. MANIM CODE GENERATION PROMPT}
\vspace{4pt}
\begin{PromptVerbatim}
Agent: CodingAgent
Purpose: Generate executable Manim code from scene plan
Output: Complete Python file with all scene classes and animations

You are an expert Manim (Community Edition) developer specializing in physics
visualizations. Generate executable Manim code implementing animations as
specified in the scene plan below. The code should create a video that explains
the physics solution in a clear, visually appealing way.

=== MANIM REFERENCE GUIDE ===
Use the following Manim classes and patterns for physics visualizations:

CORE ANIMATION CLASSES:
- Write, Create, FadeIn, FadeOut: Basic animations
- Transform, ReplacementTransform: Morphing objects
- Indicate, Circumscribe: Highlighting
- MoveAlongPath, Rotate: Motion and transformations

ESSENTIAL MOBJECTS:
- Text, MathTex: Text and equations (use MathTex for formulas!)
- Arrow, Vector, Line: Vectors and paths
- Circle, Dot, Rectangle, Polygon: Shapes
- Axes, NumberPlane: Coordinate systems
- VGroup: Group related objects

POSITIONING:
- .next_to(obj, direction, buff=0.25)
- .to_edge(direction), .to_corner()
- .shift(), .move_to(), .align_to()
- .arrange(direction, buff=0.25)

PHYSICS PATTERNS:
```python
# Force vector
force = Arrow(ORIGIN, 2*RIGHT, color=RED, buff=0)
label = MathTex("F").next_to(force, UP)

# Motion path
path = Line(LEFT*3, RIGHT*3)
obj = Dot().move_to(path.get_start())
self.play(MoveAlongPath(obj, path))
```

MANIM-PHYSICS PLUGIN (optional):
- Available for advanced physics (pendulums, fields)
- Import: from manim_physics import *
- Use only if needed for complex simulations

PROBLEM INFORMATION:
Question: {question}
Solution: {solution}

SCENE PLAN:
{scene_plan}

SPECIFIC       REQUIREMENTS            AND          RESTRICTIONS:
1. Use Manim Community Edition syntax.
2. Create a single Python file with all necessary classes and imports.
3. CRITICAL - VOICEOVER SETUP (MUST BE EXACT):
```python
from manim import *
from manim_voiceover import VoiceoverScene
import sys
import os
# CRITICAL: Add project root to Python path (4 levels up from code directory)
current_dir     = os.path.dirname(os.path.abspath(__file__))
project_root     = os.path.abspath(os.path.join(current_dir, '../../../..'))
if project_root not in sys.path:
    sys.path.insert(0, project_root)
from utils.kokoro_voiceover import KokoroService
```
4. Each scene should be a separate class inheriting from VoiceoverScene.
5. In EVERY scene's construct() method, FIRST LINE must initialize speech service:
   self.set_speech_service(KokoroService(voice="af_bella", speed=1.0, lang="en-us"))
6. Use self.voiceover(text="...") context manager for narration synchronized with animations.
7. IMPORTANT NARRATION STYLE:
   - First scene should start with a teacher-like introduction: "Today, we will learn about [topic]..."
     or "Hello! Today we're going to understand..."
   - Last scene should end with a teacher-like conclusion: "I hope you understand the solution now." or
     "That's how we solve this problem. I hope this was clear!"
   - Use warm, encouraging teaching language throughout
8. Include detailed comments explaining the code.
9. Use MathTex for all mathematical equations and formulas for better rendering.
10. DO NOT use ImageMobject or load any external images.
11. Use ONLY geometric shapes, lines, Text, MathTex, and other built-in Manim objects.
12. Ensure all animation code is complete and doesn't cut off mid-statement.
13. Create visually engaging animations for physics concepts (e.g., moving objects,
    vectors for forces, graphs for motion).

EXAMPLE MANIM CODE FORMAT:
```python
from manim import *
from manim_voiceover import VoiceoverScene
import sys
import os
# CRITICAL: Add project root to Python path to find utils module (4 levels up)
current_dir                                   = os.path.dirname(os.path.abspath(__file__))
project_root                                  = os.path.abspath(os.path.join(current_dir, '../../../..'))
if project_root not in sys.path:
    sys.path.insert(0, project_root)
from            utils.kokoro_voiceover    import            KokoroService

class IntroScene(VoiceoverScene):
   def construct(self):
      # Initialize speech service with Kokoro TTS - MUST BE FIRST
      self.set_speech_service(KokoroService(voice="af_bella", speed=1.0, lang="en-us"))

      # Title
      title = Text("Newton's Second Law")
      with self.voiceover(text="Hello! Today we will learn about kinematics and how to solve motion problems.") as tracker:
          self.play(Write(title), run_time=tracker.duration * 0.8)
      self.wait(tracker.duration * 0.2)
      self.play(title.animate.to_edge(UP))

      # Problem Text
      question_text = Text("A car accelerates uniformly from rest...", font_size=24)
      with self.voiceover(text="A car starts from rest and accelerates uniformly.") as tracker:
          self.play(Write(question_text), run_time=tracker.duration)

      # Diagram
      car = Square(side_length=0.5, color=BLUE)
      with self.voiceover(text="We can visualize this motion.") as tracker:
          self.play(Create(car))
          self.play(car.animate.shift(RIGHT*3), run_time=tracker.duration * 0.8)

class Scene2_Calculations(VoiceoverScene):
    def construct(self):
        # Initialize speech service with Kokoro TTS
        self.set_speech_service(KokoroService(voice="af_bella", speed=1.0, lang="en-us"))

        # Equations using MathTex
        equation = MathTex("v = u + at", font_size=36)
        with self.voiceover(text="We use the equation v equals u plus a t.") as tracker:
            self.play(Write(equation), run_time=tracker.duration)

if __name__ == "__main__":
    # This will be handled by the system
    pass
```

IMPORTANT:
- Your code must be complete, executable, and error-free.
- Use MathTex for equations.
- FOCUS on creating clear physics visualizations.
- IMPLEMENT all scenes from the scene plan.
[Note: If RAG documentation is available, it is appended here]
\end{PromptVerbatim}

\vspace{8pt}
\subsubsection*{4. MANIM CODE ERROR FIXING PROMPT}
\vspace{4pt}
\begin{PromptVerbatim}
Agent: CodingAgent
Purpose: Debug and fix Manim code that failed to execute
Output: Corrected Python code with fixes applied

You are an expert Manim (Community Edition) developer specializing in debugging
physics animations. The following Manim code failed to execute. Analyze the error
message and the code, then provide a corrected version.

SCENE PLAN:
{scene_plan}

FAILED CODE:
```python
{code}
```

ERROR MESSAGE:
{error_message}

SPECIFIC REQUIREMENTS FOR FIXING:
1. Identify the cause of the error.
2. Correct the code to resolve the error.
3. Ensure the corrected code still meets the original scene requirements.
4. Use Manim Community Edition syntax.
5. Use MathTex for all mathematical equations.
6. CRITICAL - If ModuleNotFoundError for 'utils', add this EXACT code at the top
   after imports:
   ```python
   from manim import *
   from manim_voiceover import VoiceoverScene
   import sys
   import os
   # CRITICAL: Add project root to Python path (4 levels up from code directory)
   current_dir = os.path.dirname(os.path.abspath(__file__))
   project_root = os.path.abspath(os.path.join(current_dir, '../../../..'))
   if project_root not in sys.path:
       sys.path.insert(0, project_root)
   from utils.kokoro_voiceover import KokoroService
   ```
7. Ensure EVERY scene class has this as FIRST LINE in construct():
   self.set_speech_service(KokoroService(voice="af_bella", speed=1.0, lang="en-us"))
8. DO NOT create mock/fake KokoroService classes - use the real one from utils.kokoro_voiceover
9. DO NOT remove manim_voiceover imports - VoiceoverScene is required for audio
10. Return only the full, corrected Python code. Do not include explanations outside the code comments.

Corrected Code:
[Note: If RAG documentation is available, it is appended here]
\end{PromptVerbatim}

\vspace{8pt}
\subsubsection*{5. SCREENSHOT FEEDBACK IMPROVEMENT PROMPT}
\vspace{4pt}
\begin{PromptVerbatim}
Agent: CodingAgent
Purpose: Improve Manim code based on visual quality feedback from screenshots
Output: Enhanced code with better layout, spacing, and visual clarity

You are an expert Manim (Community Edition) developer specializing in visual
quality and layout optimization for physics animations. The following Manim code
has been rendered and screenshots were analyzed. Based on the visual feedback from
the screenshot analysis, improve the code to address layout, spacing, overlap, and
visibility issues.

SCENE PLAN:
{scene_plan}

CURRENT CODE:
```python
{code}
```

VISUAL FEEDBACK FROM SCREENSHOT ANALYSIS:
{visual_feedback}

SPECIFIC REQUIREMENTS FOR VISUAL IMPROVEMENTS:
1. **Overlap Issues**: If text or objects overlap, adjust positions, font sizes,
   or use `arrange()` and `next_to()` to create proper spacing.
2. **Spacing and Layout**: Ensure equations, text, and diagrams have clear visual
   hierarchy and sufficient whitespace.
3. **Font Size**: Increase or decrease font_size for Text and MathTex objects to
   improve readability.
4. **Positioning**: Use `to_edge()`, `shift()`, `next_to()`, `arrange()`, and
   alignment constants (UP, DOWN, LEFT, RIGHT) to optimize object placement.
5. **Color and Contrast**: Ensure colors provide good contrast and visibility
   (e.g., avoid light colors on white backgrounds).
6. **Animation Timing**: Adjust wait times if elements appear too quickly or slowly.
7. **Visual Clarity**: Simplify overly complex scenes by breaking them into smaller
   parts or removing unnecessary elements.
8. **Scene Relevance**: Ensure all visual elements support the narration and learning
   objectives.

IMPORTANT GUIDELINES:
- Use Manim Community Edition syntax only.
- Use MathTex for all mathematical equations.
- Ensure KokoroService is initialized:
  self.set_speech_service(KokoroService(voice="af_bella", speed=1.0, lang="en-us"))
- DO NOT use ImageMobject or load external images.
- Maintain the original scene structure and content while improving layout and spacing.
- Include detailed comments explaining the visual improvements made.
- Return only the full, improved Python code. Do not include explanations outside
  the code comments.

EXAMPLE IMPROVEMENTS:
- If "Scene1: high overlap detected" → Add `.arrange(DOWN, buff=0.5)` or use
  `next_to()` with appropriate buffers
- If "text too small" → Increase `font_size=36` to `font_size=48`
- If "equation off-screen" → Use `.to_edge(UP)` or `.shift(UP*0.5)`
- If "elements clustered" → Use `VGroup()` with `.arrange()` to distribute elements evenly

Improved Code:
\end{PromptVerbatim}

\vspace{8pt}
\subsubsection*{6. SOLUTION EVALUATION PROMPT (LLM-as-Judge)}
\vspace{4pt}
\begin{PromptVerbatim}
Agent: SolutionEvaluator
Purpose: Evaluate correctness of physics solution using research-grade rubrics
Output: JSON with 5 sub-scores (0-5 scale) and overall score

You are an expert physics professor evaluating a solution for research purposes.
Use the explicit rubrics below to score the solution on a 0-5 scale for each metric.
Be objective, deterministic, and justify your scores with evidence.

QUESTION:
{question}

SOLUTION:
{solution}

EVALUATION RUBRICS (0-5 scale):
1. EQUATION CORRECTNESS (0-5):
   5: All equations correct, properly derived, LaTeX formatted
   4: Minor notation issues or missing intermediate steps
   3: 1-2 major equation errors but approach correct
   2: Multiple equation errors, flawed approach
   1: Fundamentally wrong equations or missing
   0: No equations or completely incorrect

2. NUMERICAL ACCURACY (0-5):
   5: All numerical answers correct within 1% tolerance
   4: Correct within 5% tolerance
   3: Correct method, minor calculation errors
   2: Correct approach, significant calculation errors
   1: Wrong final answers due to conceptual errors
   0: Completely incorrect numerical results

3. STEP COMPLETENESS (0-5):
   5: All required steps present and explained
   4: Missing 1 minor step
   3: Missing 1 major step but derivable
   2: Missing 2+ major steps
   1: Solution jumps to answer without proper steps
   0: No steps shown, just final answer

4. PHYSICS CONCEPT COVERAGE (0-5):
   5: All relevant concepts explicitly mentioned and applied
   4: 1 minor concept implied but not stated
   3: 1 major concept missing or incorrect
   2: 2+ concepts missing
   1: Fundamental concepts misunderstood
   0: No relevant physics concepts mentioned

5. MATHEMATICAL RIGOR (0-5):
   5: Proper notation, clear derivations
   4: Minor notation inconsistencies
   3: Some unclear or poorly justified steps
   2: Logical gaps in derivation
   1: Mathematically unsound reasoning
   0: No mathematical justification

Provide your evaluation in the following strict JSON format:
{
   "equation_correctness": <0-5>,
   "equation_correctness_evidence": "Brief justification with specific examples",
   "numerical_accuracy": <0-5>,
   "numerical_accuracy_evidence": "Brief justification with specific examples",
   "step_completeness": <0-5>,
   "step_completness_evidence": "Brief justification with specific examples",
   "concept_coverage": <0-5>,
   "concept_coverage_evidence": "Brief justification with specific examples",
   "mathematical_rigor": <0-5>,
   "mathematical_rigor_evidence": "Brief justification with specific examples",
   "overall_score": <average of above 5 metrics>,
   "strengths": ["specific strength 1", "specific strength 2"],
   "weaknesses": ["specific weakness 1", "specific weakness 2"],
   "confidence": <0-5, your confidence in this evaluation>,
   "feedback": "Concise summary of evaluation"
}

Return ONLY valid JSON. No additional text.
\end{PromptVerbatim}

\vspace{8pt}
\subsubsection*{7. EXPLANATION EVALUATION PROMPT (LLM-as-Judge)}
\vspace{4pt}
\begin{PromptVerbatim}
Agent: ExplanationEvaluator
Purpose: Evaluate pedagogical quality of video explanation plan
Output: JSON with 5 sub-scores (0-5 scale) and overall score

You are an expert in physics pedagogy and educational psychology evaluating an
explanation for research purposes. Use the explicit rubrics below to score the
explanation on a 0-5 scale for each metric.

QUESTION:
{question}

SOLUTION:
{solution}

SCENE PLAN (Explanation Breakdown):
{scene_plan}

EVALUATION RUBRICS (0-5 scale):
1. LOGICAL FLOW (0-5):
   5: Perfect progression from problem to solution, each step builds on previous
   4: Clear flow with 1 minor transition issue
   3: Generally logical but 1-2 jumps or unclear transitions
   2: Multiple logical gaps or confusing sequence
   1: Disorganized, hard to follow progression
   0: No logical flow, completely scattered

2. PEDAGOGICAL CLARITY (0-5):
   5: Concepts explained at appropriate level, clear language, no ambiguity
   4: Very clear with 1 minor terminology/explanation issue
   3: Mostly clear but some jargon or unexplained terms
   2: Several unclear explanations or confusing statements
   1: Unclear, uses unexplained jargon, assumes too much knowledge
   0: Completely unclear or incomprehensible

3. VISUALIZATION ALIGNMENT (0-5):
   5: Each scene has clear visual purpose, animations match explanation perfectly
   4: Strong alignment with 1 minor mismatch
   3: Good alignment but some scenes lack clear visual purpose
   2: Several mismatches between visuals and explanation
   1: Poor alignment, visuals do not support understanding
   0: No alignment between visuals and explanation

4. INTUITION BUILDING (0-5):
   5: Builds deep understanding, connects to prior knowledge, provides insights
   4: Good intuition building with 1 missed opportunity
   3: Some intuition provided but could be deeper
   2: Limited intuition, mostly procedural
   1: No intuition building, purely mechanical
   0: Completely procedural, no understanding fostered

5. PACING & ACCESSIBILITY (0-5):
   5: Perfect pacing for target audience, appropriate difficulty progression
   4: Good pacing with 1 scene too fast/slow
   3: Generally appropriate but some pacing issues
   2: Multiple pacing problems, difficulty jumps
   1: Poor pacing, inappropriate for audience level
   0: Completely inappropriate pacing or difficulty

Provide your evaluation in the following strict JSON format:
{
  "logical_flow": <0-5>,
  "logical_flow_evidence": "Brief justification",
  "pedagogical_clarity": <0-5>,
  "pedagogical_clarity_evidence": "Brief justification",
  "visualization_alignment": <0-5>,
  "visualization_alignment_evidence": "Brief justification",
  "intuition_building": <0-5>,
  "intuition_building_evidence": "Brief justification",
  "pacing_accessibility": <0-5>,
  "pacing_accessibility_evidence": "Brief justification",
  "overall_score": <average of above 5 metrics>,
  "strengths": ["specific strength 1", "specific strength 2"],
  "weaknesses": ["specific weakness 1", "specific weakness 2"],
  "suggestions": ["specific suggestion 1", "specific suggestion 2"],
  "confidence": <0-5>,
  "feedback": "Concise summary"
}

Return ONLY valid JSON. No additional text.
\end{PromptVerbatim}

\vspace{8pt}
\subsubsection*{8. SCREENSHOT VISUAL QUALITY ANALYSIS PROMPT (Vision LLM)}
\vspace{4pt}
\begin{PromptVerbatim}
Agent: ScreenshotAnalyzer
Purpose: Analyze video screenshot visual quality using Vision LLM
Output: JSON with 5 visual quality scores (0-5 scale) and evidence

You are an expert in educational video production analyzing a Manim animation
screenshot for research purposes.

** CRITICAL FIRST STEP - SCENE COMPLETENESS CHECK:
Before evaluating quality, determine if this screenshot shows a COMPLETE, EVALUABLE scene:

EVALUABLE (proceed with scoring):
- Scene has substantial content: multiple text elements, equations, diagrams, or animations
- Educational content is clearly visible and developed
- Scene appears to be at a meaningful state (not just starting/ending)
- Enough visual elements present to assess layout, readability, and alignment

NOT EVALUABLE (skip scoring, mark as incomplete):
- Nearly blank screen with only 1-2 lines of text appearing/disappearing
- Scene in transition with minimal content visible
- Just a title or single element on screen
- Scene clearly caught mid-animation before content fully appears

IF NOT EVALUABLE: Return JSON with "evaluable": false, "reason": "<why not evaluable>",
and set all scores to null.
IF EVALUABLE: Proceed with strict evaluation below.

IMPORTANT SCORING GUIDELINES (for evaluable scenes):
- Use WHOLE NUMBERS ONLY (0, 1, 2, 3, 4, or 5) for individual metrics
- Be very stringent - most animations will have issues
- Look carefully for overlapping elements, text cut-offs, LaTeX rendering errors
- Only give 5 if the screenshot is nearly perfect
- Give 4 for good quality with minor issues
- Give 3 for noticeable problems
- Give 2 for significant issues
- Give 0-1 for serious/severe issues

SCENE {scene_index} INFORMATION:
{scene_data}

SCREENSHOT TIMESTAMP:
{timestamp_label}

VISUAL QUALITY RUBRICS (0-5 scale, WHOLE NUMBERS ONLY - BE STRICT, only for EVALUABLE scenes):
CRITICAL: Check for these common issues and penalize heavily:
- Overlapping text or equations (reduce layout_quality by 1-2 points)
- LaTeX rendering errors like "\\text" or broken symbols (reduce equation_rendering by 1-2 points)
- Any content cut off at screen edges (reduce off_screen_issues by 1-2 points)
- Text too small to read comfortably (reduce text_readability by 1-2 points)
- Misaligned equations or text (reduce layout_quality by 1 point)

1. LAYOUT QUALITY (0-5 whole number):
   5: PERFECT spacing, hierarchy, balance; zero crowding, zero overlap, zero empty space issues
   4: Excellent layout with only 1 very minor spacing inconsistency
   3: Good layout but has minor positioning issues or slight overlap
   2: Noticeable spacing/positioning problems, some elements overlap
   1: Multiple overlapping elements, poor spacing, cluttered appearance
   0: Severe overlap, completely cluttered, unusable layout

2. TEXT READABILITY (0-5 whole number):
   5: ALL text perfectly legible, optimal font sizes (not too small/large), excellent contrast
   4: Very readable, one piece of text slightly suboptimal
   3: Mostly readable but some text is too small, too large, or low contrast
   2: Several readability issues, some text hard to read
   1: Multiple text elements illegible or poorly sized
   0: Most text illegible, missing, or completely unreadable

3. EQUATION RENDERING (0-5 whole number):
   5: ALL LaTeX equations perfectly rendered, no errors, perfect alignment and spacing
   4: Excellent rendering with 1 very minor formatting inconsistency
   3: Good but has some misalignment, spacing issues, or minor LaTeX errors
   2: Multiple equation rendering problems, visible LaTeX errors, poor formatting
   1: Serious LaTeX rendering errors, equations malformed or misaligned
   0: Equations completely broken, unreadable, or incorrectly rendered

4. OFF-SCREEN ISSUES (0-5 whole number):
   5: ALL elements fully visible, perfect framing, nothing cut off
   4: One very minor, non-critical element barely touching edge
   3: Some minor elements slightly cut off at edges
   2: Important elements partially off-screen or cut off
   1: Critical content cut off, multiple elements off-screen
   0: Major content missing due to being off-screen

5. SCENE-CONTENT ALIGNMENT (0-5 whole number):
   5: Visuals PERFECTLY match intended content, narration, and scene description
   4: Strong alignment with only 1 very minor mismatch
   3: Good alignment but some visual elements don't fully match description
   2: Several mismatches between visuals and intended content
   1: Significant mismatches, visuals don't represent content well
   0: Visuals completely don't match intended scene content

Provide your analysis in the following strict JSON format:

IF NOT EVALUABLE:
{
   "evaluable": false,
   "reason": "Specific reason why scene is incomplete/not evaluable",
   "layout_quality": null,
   "text_readability": null,
   "equation_rendering": null,
   "off_screen_issues": null,
   "scene_content_alignment": null,
   "visual_quality_score": null,
   "confidence": 0
}

IF EVALUABLE:
{
   "evaluable": true,
   "layout_quality": <whole number 0-5>,
   "layout_quality_evidence": "Specific observations about spacing, overlap, positioning",
   "text_readability": <whole number 0-5>,
   "text_readability_evidence": "Specific observations about font sizes, contrast, legibility",
   "equation_rendering": <whole number 0-5>,
   "equation_rendering_evidence": "Specific observations about LaTeX rendering, errors, alignment",
   "off_screen_issues": <whole number 0-5>,
   "off_screen_issues_evidence": "Specific observations about elements cut off or off-screen",
   "scene_content_alignment": <whole number 0-5>,
   "scene_content_alignment_evidence": "Specific observations about visual-content match",
   "issues": ["specific issue 1 with severity", "specific issue 2 with severity"],
   "suggestions": ["specific actionable suggestion 1", "specific actionable suggestion 2"],
   "confidence": <whole number 0-5>,
   "feedback": "Concise critical summary highlighting all problems found"
}

NOTE: Do NOT include "visual_quality_score" in your response. It will be calculated automatically
as the average of the LOWEST 2 scores from the 5 metrics above (to avoid inflation from
non-applicable metrics).

Return ONLY valid JSON. No additional text.
\end{PromptVerbatim}

\end{document}